\documentclass{article}

\usepackage{wrapfig}

\usepackage[numbers]{natbib}
\usepackage[final]{neurips_2023}

\usepackage[utf8]{inputenc} 
\usepackage[T1]{fontenc}    
\usepackage{url}            
\usepackage{booktabs}       
\usepackage{amsfonts}       
\usepackage{nicefrac}       
\usepackage{microtype}      
\usepackage{xcolor}         
\usepackage{amsmath}
\usepackage{graphicx}
\usepackage[noline,ruled]{algorithm2e}

\title{DGFN: Double Generative Flow Networks}

\author{
  Elaine Lau\textsuperscript{1,2}, Nikhil Vemgal\textsuperscript{1,2}, Doina Precup\textsuperscript{1,2,3}, Emmanuel Bengio\textsuperscript{4}\\
  \textsuperscript{1}Mila, \textsuperscript{2}McGill University, \\
  \textsuperscript{3}Google DeepMind, \textsuperscript{4} Valence Labs \\
  \texttt{\{tsoi.lau, nikhil.vemgal\}@mail.mcgill.ca} \\
  \texttt{dprecup@cs.mcgill.ca, bengioe@gmail.com}
}

\pdfoutput=1
\begin{document}

\maketitle

\begin{abstract}


Deep learning is emerging as an effective tool in drug discovery, with potential applications in both predictive and generative models. Generative Flow Networks (GFlowNets/GFNs) are a recently introduced method recognized for the ability to generate diverse candidates, in particular in small molecule generation tasks. In this work, we introduce double GFlowNets (DGFNs). Drawing inspiration from reinforcement learning and Double Deep Q-Learning, we introduce a target network used to sample trajectories, while updating the main network with these sampled trajectories. Empirical results confirm that DGFNs effectively enhance exploration in sparse reward domains and high-dimensional state spaces, both challenging aspects of de-novo design in drug discovery.

\end{abstract}

\section{Introduction}

One of the greatest challenges in modern medicine currently lies in the discovery and development of novel therapeutics for disease treatment. This challenge is most evident in the field of infectious diseases, where the creation of new antibiotics has been challenging due to substantial research costs, lengthy timelines, and limited returns. In recent years, a promising alternative approach has emerged in the form of Generative Flow Networks (GFlowNets) \cite{GFN_FM2021, GFN_foundation_2021}. GFlowNets tackle the sampling problem by learning to sample trajectories approximately proportionally to their quality, as captured by a  reward function. 
This approach encourages the discovery of a diverse set of high-reward samples, offering the potential to significantly accelerate the drug discovery and development process.

Despite existing research on credit assignment for GFlowNets, a central challenge continues to be the improvement of exploration and exploitation within the GFlowNets framework \cite{GFN_TB2022, GFN_subTB2022, GFN_Thompson, GFN_Bayesian}. 
In environments like small molecule generation, where the rewards are sparse \cite{GFN_FM2021}, GFlowNets may encounter difficulties in breaking away from the current best mode, thus reducing the chances for the agent to discover new modes in the environment \cite{GFN_GAFN, GFN_foundation_2021}.
Therefore, it is important to find new ways to enhance exploration efficiency in sparse-reward domains.

In this work, we take inspiration from the double deep Q-learning (DDQN) algorithm from reinforcement learning \cite{DDQN2015}, and introduce double GFlowNets (DGFNs). 
This approach simply involves employing a target network, which acts as the delayed version of the online network and from which we generate trajectories. Intuitively, this prevents the data distribution on which the online model is trained to become too peaked, too ``opinionated", too quickly. We apply DGFNs to two standard GFN tasks: hypergrid (where the complexity and sparsity can be controlled well) and small molecule generation (which is more illustrative of real applications)~\citep{GFN_FM2021}. Our empirical findings demonstrate that DGFNs finds all modes faster in hypegrid, and uncover a greater number of high-reward modes in the fragment-based molecular design task. These observations provide strong evidence that our proposed strategy indeed encourages diverse exploration, promoting better coverage of the state space and the discovery of diverse candidate solutions. 




\section{Preliminaries}

We begin by introducing GFlowNets, following previous work~\cite{GFN_FM2021, GFN_TB2022}. 
Consider a directed acyclic graph $G=(S,A)$, where each vertex $s \in S$ represents a state and $s\to s'\in A$ a state transition. Notably, $s_0$ is the initial state, with no incoming edges, while $s_f$ is the sink state, with no outgoing edges. A state $s_n$ is considered terminal if $s_n\to s_f\in A$. Each state is assumed to represent some object $s_n=x\in\mathcal{X}$. We sample such objects by sampling trajectories starting from $s_0$, and following ``actions" $a$ drawn from $A$. This yields a sequence $\tau = (s_0 \rightarrow s_1 \rightarrow \ldots s_n \rightarrow s_f)$, referred to as a \textit{complete trajectory}. Let $\mathcal{T}$ be the set of all possible trajectories. Assuming a probability distribution over the edges from each node, let $F(\tau)$ denote the flow of $\tau$, representing its unnormalized probability. The  \textit{edge flow} is defined as $F\left(s \rightarrow s^{\prime}\right) = \sum_{\tau \in \mathcal{T}:(s \rightarrow s^{\prime}) \in \tau} F(\tau)$. The \textit{state flow} is defined as $F(s) = \sum_{\tau\in\mathcal{T}:s \in \tau} F(\tau)$. Using these concepts, we can define forward and backward policies, $P_F$ and $P_B$, as follows::
\begin{equation}
    P_F(s\to s') = \frac{F(s\to s')}{F(s)}, \;\; P_B(s'\to s)= \frac{F(s\to s')}{F(s')}
\end{equation}
Given a non-negative reward function $R: \mathcal{X} \rightarrow \mathbb{R}_{\geq 0}$, let the terminal edge flows be $F(s_n\to s_f) = R(x=s_n)$. The primary objective of GFlowNets is to train a generative policy such that the likelihood of sampling $x \in \mathcal{X}$ is proportional to $R(x)$ 
, where
\begin{equation}
p(x) =\sum_{\tau=\left(s_0 \rightarrow \cdots \rightarrow s_n=x\right)} P_F(\tau)=\sum_\tau \prod_{t=1}^n P_F\left(s_t \mid s_{t-1}\right)
\end{equation}
This is achieved by \emph{balancing} flows such that the total quantity of flow is preserved. \cite{GFN_FM2021} In particular, this can be expressed through the 
\textit{trajectory balance} (TB) \cite{GFN_TB2022} condition, where for all trajectories $\tau$:
\begin{equation}
\label{eqn:TB}
Z \prod_{t=1}^n P_F\left(s_t \mid s_{t-1}\right)=R(x) \prod_{t=1}^n P_B\left(s_{t-1} \mid s_t\right)
\end{equation}
Here, $Z$ represents the total flow, i.e. $Z=\sum_{x\in\mathcal{X}}R(x)$. This equality can be turned into an objective and used to learn parameterized $P_F$, $P_B$, and $Z$~\citep{GFN_TB2022, GFN_subTB2022}. Satisfying this constraint across all complete trajectories ensures that $P_F(x) \propto R(x)$. Throughout our experiments, we adopt trajectory balance as the primary training objective.

\subsection{Related Work}

\textbf{Double Deep Q-Learning (DDQN)}
In reinforcement learning, it is well-known that the Q-learning algorithm for learning optimal policies suffers from overestimation bias~\cite{Double_qlearning, double_qlearning2010}. This overestimation leads to collapsing the exploration too quickly, slowing down the learning process. Double Deep Q-Networks (DDQN) were introduced to mitigate this issue, by decoupling action selection and value estimation \cite{Double_qlearning}. DDQN uses two separate Q-networks: a target network and an online network. The target network is used to compute the Q-learning target, which is used to update the weights of the online value network. The latter is employed for action selection. The target network is updated periodically by copying the weights of the online network. DDQN has been shown empirically to provide more stable Q-value estimates, thereby enhancing performance.

\textbf{Improving GFlowNets} A number of works have delved into enhancing the training process by manipulating the sampling distribution. For example, Thompson Sampling GFlowNets \cite{GFN_Thompson}, improve exploration by maintaining uncertainty through ensembling and using it within Thompson sampling. Replay buffers coupled with reward-prioritized replay sampling \cite{GFN_concurrentPER, GFN_PER} have been shown to enhance GFlowNet training dynamics. However, existing work only considers using single networks to generate trajectories.  

\section{Double Generative Flow Networks (DGFN)}

In environments characterized by large state spaces or sparse reward signals, the standard approach to GFlowNet training can induce training instability and/or lead to mode collapse~\citep{GFN_concurrentPER}. To address this limitation and promote exploration, we draw inspiration from the DDQN idea. 
We introduce a target network, whose primary role is to sample trajectories, on which the the online network's loss is computed. In our experiments,  this approach safeguards against over-optimization of the GFlowNet, particularly in environments characterized by sparse rewards, where the agent may be pushed to exploit known modes if it is able to learn about them quickly,  thereby exploring too little. The target network dampens this problem. 
The target network is periodically copied from the online network, but this process is a bit different from standard DDQN. Because the initial sampled trajectories tend to be relatively random and low reward, 
 in the initial stages of training, we update the target network more frequently. Subsequently, we transition to periodic updates employing a Polyak averaging technique.
A full description of the approach is given in Algorithm 1.

\DontPrintSemicolon
\begin{algorithm}[H]
 \caption{Double GFlowNets (DGFNs)}
  \KwIn{Initial phase length $T^I$, update period $T^U$} 
  Initialize online flow network $F_{\theta}$, target flow network $F_{\theta'}$, $\alpha < 1$ \;
    \For{each training step $t=1$ to $T$}
    {
      Sample a batch of $M$ trajectories $\tau=\left\{s_0 \rightarrow \cdots \rightarrow s_n\right\}$ from $F_{\theta'}$ \;

      Compute loss of the online network using sampled trajectories \;
        
        \uIf{$t< T^I \textup{ or } t\bmod T^U \equiv 0$}{%
          $\theta^{\prime} \leftarrow \alpha  \theta+(1-\alpha)  \theta^{\prime}$ \;
        }

        
    }
\end{algorithm}


        

        

\section{Experiments} 


We study the performance of the proposed DGFNs in comparison to conventional GFlowNets with different objective functions on two benchmark tasks: hypergrid and molecule generation.

\subsection{Hypergrid Environment: High Dimensional, Sparse Rewards}

In the synthetic hypergrid environment introduced by Bengio et al.~\cite{GFN_FM2021}, the goal is to sample trajectories in a $D$-dimensional grid-world with side length $H$. The initial state is $(0,...,0)$, and actions increment a coordinate by 1 within the bounds of $D$ and $H$. The agent can terminate at any state. 

We use a 6-dimensional grid with side lengths $H = 8,10,12$. The reward function is defined as in~\cite{GFN_FM2021, GFN_TB2022}: 
$
R(x)=R_0+R_1 \prod_i \mathbb{I}\left(0.25<\left|x_i / H-0.5\right|\right)+R_2 \prod_i \mathbb{I}\left(0.3<\left|x_i / H-0.5\right|<0.4\right)$
with $0<R_0 \ll R_1<R_2$, $R_1=1 / 2, R_2=2$. We opt for a more challenging environment by setting $R_0=10^{-3}$, making exploration less rewarding for the agent. We measure the $L_1$ error between the true reward distribution and the empirical distribution over the sampled terminal states. Additionally, we track the number of modes discovered over the sampled terminal states. 

\begin{figure}
  \centering
  \includegraphics[scale=0.40]{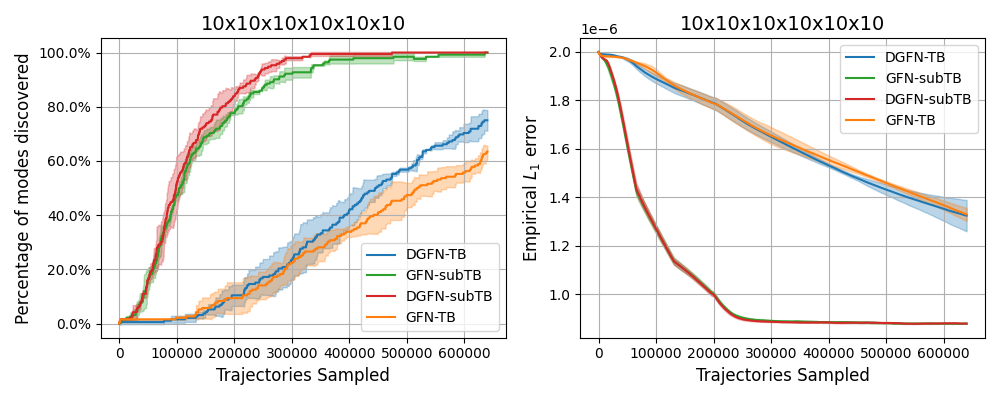}
  \caption{Left: Percentage of modes discovered over the trajectories sampled. Right: $L_{1}$ distance between empirical and target distribution over the trajectories sampled.}
   \label{fig:L1_Modes_Hypergrid}
\end{figure}

The results are shown in Fig.~\ref{fig:L1_Modes_Hypergrid} and appendix \ref{appendix:hypergrid}, with means and standard errors computed over 5 independent runs on $DGFN_{TB}$, $DGFN_{SubTB}$, and baseline models: $GFN_{TB}$ and $GFN_{SubTB}$. ``TB" refers to the objective mentioned in equation \ref{eqn:TB}, while ``SubTB" is a recent objective proposed by Madan et al. \cite{GFN_subTB2022} to learn from partial trajectories rather than complete trajectories. 
As hypothesized, the DGFNs find modes across the hypergrid faster than conventional GFlowNet methods. In particular, the gap becomes wider in more complex environments. 

\subsection{Small molecule synthesis}

To assess the capabilities of DGFNs for drug discovery, we conduct experiments involving the generation of small molecules. Specifically, this task involves creating molecules with low binding energy to the soluble epoxide hydrolase (sEH) protein, employing a docking prediction approach originally introduced by Trott and Olson~\cite{Trott_docking}. The agent is tasked with a sequential decision-making process, determining attachment points for molecular building blocks,  while adhering to the constraints of chemical validity. 
This task is particularly challenging, primarily due to the vast state space, estimated at around $10^{16}$ distinct states, and number of available actions at each state, which can range from $100$ to $2000$.
The reward function $R$ relies on a pre-trained proxy model developed by Bengio et al.~\cite{GFN_FM2021}. For implementation details, please refer to Appendix \ref{appendix:hypergrid}.

\begin{figure}
  \centering
  \includegraphics[scale=0.40]{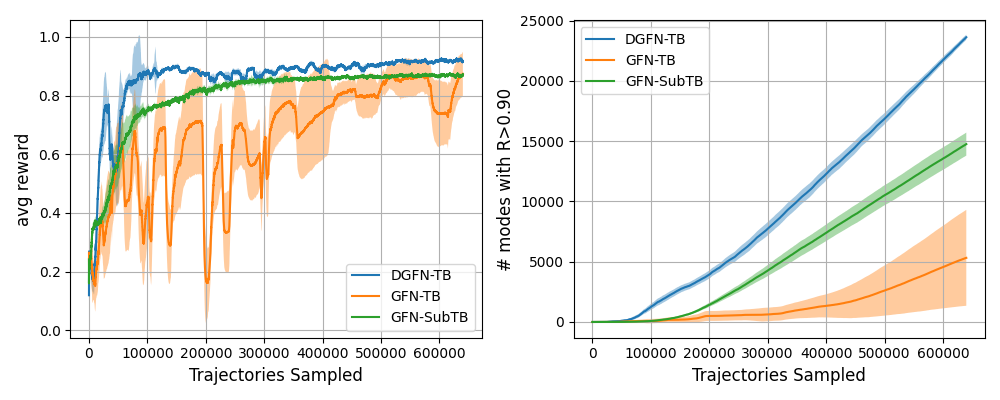}
  \caption{Left: Average reward as a function of trajectories sampled. Right: Number of modes with rewards $R>0.90$ as a function of trajectories sampled.}
  \label{fig:moles_env_rewards}
\end{figure}

We train three models: $DGFN_{TB}$ and two baseline models, $GFN_{TB}$ and $GFN_{SubTB}$. To ensure the reliability of our findings, we report reward means and standard error over 5 independent random seeds. We calculate the number of modes with rewards greater than a $0.9$ threshold.
Fig.~\ref{fig:moles_env_rewards} shows that $GFN_{TB}$ exhibits more pronounced fluctuations during the training process, whereas $DGFN_{TB}$ has lower  variance. Additionally, $DGFN_{TB}$ surpasses $GFN_{SubTB}$ in the discovery of modes with rewards exceeding $0.9$.

We also compute the diverse Top-K (i.e. the set of top molecules such that their pairwise Tanimoto similarity is at most 0.7, c.f. \citep{GFN_FM2021}) and Top-K Reward for the generated molecules, showed in Table~\ref{mol-table}. Here, $DGFN_{TB}$ matches the baselines. This shows that $DGFN_{TB}$ is able to find a greater diversity of solutions without sacrificing reward. Appendix \ref{appendix:molesynthesis} provides examples of the top 12 molecules generated by our model.

\begin{table}
  \caption{Results on the molecule synthesis task. Mean and standard error over 3 runs.}
  \label{mol-table}
  \centering
  \begin{tabular}{ccccc}
    \toprule

Algorithm & Diverse Top-100 & Diverse Top-1000 & Top-100 Reward & Top-1000 Reward \\ \midrule
DGFN-TB   & \textbf{1.035$\pm$0.002}       & \textbf{1.017$\pm$0.002}        & \textbf{1.036$\pm$0.002}    & \textbf{1.020$\pm$0.002}    \\
GFN-TB    & 0.972$\pm$0.028       & 0.836$\pm$0.097        & 0.982$\pm$0.024    & 0.931$\pm$0.044    \\
GFN-SubTB   & 1.017$\pm$0.001       & 0.992$\pm$0.002        & 1.017$\pm$0.001    & 0.996$\pm$0.002 \\
    \bottomrule
  \end{tabular}
\end{table}

\section{Discussion and Conclusion}

In this work, we introduced the concept of Double Generative Flow Networks (DGFNs), in order to improve the training stability and the ability of GFlowNets to explore well in large state spaces with sparse rewards. This issue is especially important in drug discovery, where sampling molecules demands a more robust exploration strategy.  Our empirical results in hypergrid and molecule synthesis tasks demonstrate the effectiveness of DGFN in promoting diversity in sample generation and enhancing stability. More work remains: more extensive testing of DGFNs across different tasks, the development of a more comprehensive theoretical framework for this approach, and ultimately,  the exploration of more techniques inspired by RL and generative modeling to improve the stability of GFlowNets.

\section*{Acknowledgements}
We genuinely appreciate the funding support from Fonds Recherche Quebec through the FACS-Acquity grant and the National Research Council of Canada. Mila has been instrumental in providing the computational resources for this project. We also want to acknowledge Jarrid Rector Brooks and Moksh Jain for their valuable discussions on related work. This work is partially done at Valence Labs. 

{\small \bibliography{neurips_2023}}
\bibliographystyle{plain}

\newpage
\appendix
\section{Appendix}

\subsection{Experiment details: Hypergrid}\label{appendix:hypergrid}


 The model architecture for both the forward and backward policies remains consistent with the original GFlowNets models \cite{GFN_TB2022, GFN_subTB2022}, using Adam as the optimizer. All models were trained using a batch size of 64 for a total of 640,000 trajectories. Hyperparameter tuning was conducted via Optuna \cite{Optuna}, which automates hyperparameter optimization with Ray Tune \cite{Ray}. For dimension 6, side length 10, the optimal hyperparameters for $DGFN_{TB}$ were found to be an initial phase length of $T^I = 698$ and an update period of $T^U = 137$. For $DGFN_{SubTB}$, the optimal settings were an initial phase length of $T^I = 794$ and an update period of $T^U = 149$.

\begin{figure}[h]
  \centering
  \includegraphics[scale=0.45]{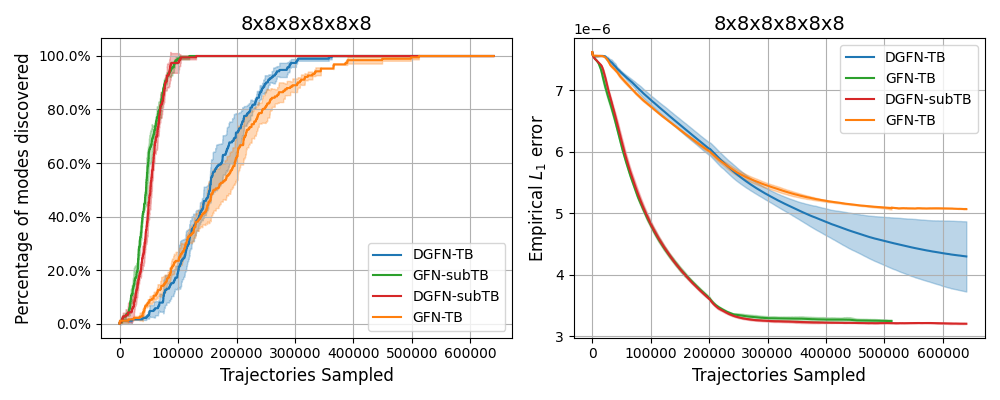}
  \includegraphics[scale=0.45]{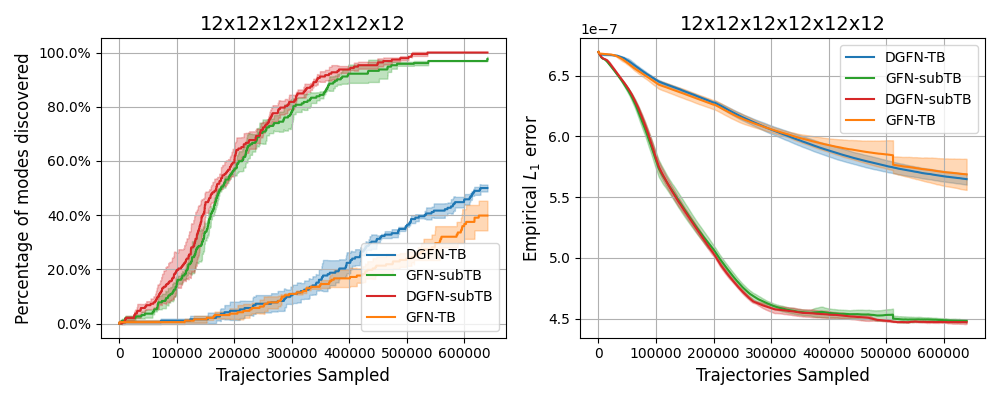}
  \caption{Left: Average reward as a function of trajectories sampled. Right: Number of modes with
rewards R > 0.90 as a function of trajectories sampled.}
  \label{fig:appendix_grid}
\end{figure}

\subsection{Experiment details: Small Molecule Synthesis}\label{appendix:molesynthesis}

In our experiments, we used the dataset and proxy model provided by Bengio et al.~\cite{GFN_FM2021}. The model architecture has the same implementation as the GFlowNet's trajectory balance~\cite{GFN_TB2022}. Additionally, we incorporated the AutoDock Vina library~\cite{Trott_docking} for binding energy estimation and relied on the RDKit library~\cite{rdkit} for chemistry routines.

For the experimental setup, we trained the models for a total of 10,000 iterations using a batch size of 64. The temperature coefficient for the reward function, denoted as $\beta$ ~\cite{GFN_FM2021}, was set to 96. To determine the optimal initial phase length $T^I$ and update period $T^U$, we conducted a grid search over the following values: $T^I \in \{500, 1000, 1500, 2000, 2500, 3000\}$ and $T^U \in \{50, 100, 150, 200, 250, 300, 350\}$. Our experiments revealed that $T^I$ of 2000 and $T^U$ of 200 produced the best results.

\begin{figure}
  \centering
  \includegraphics[scale=0.9]{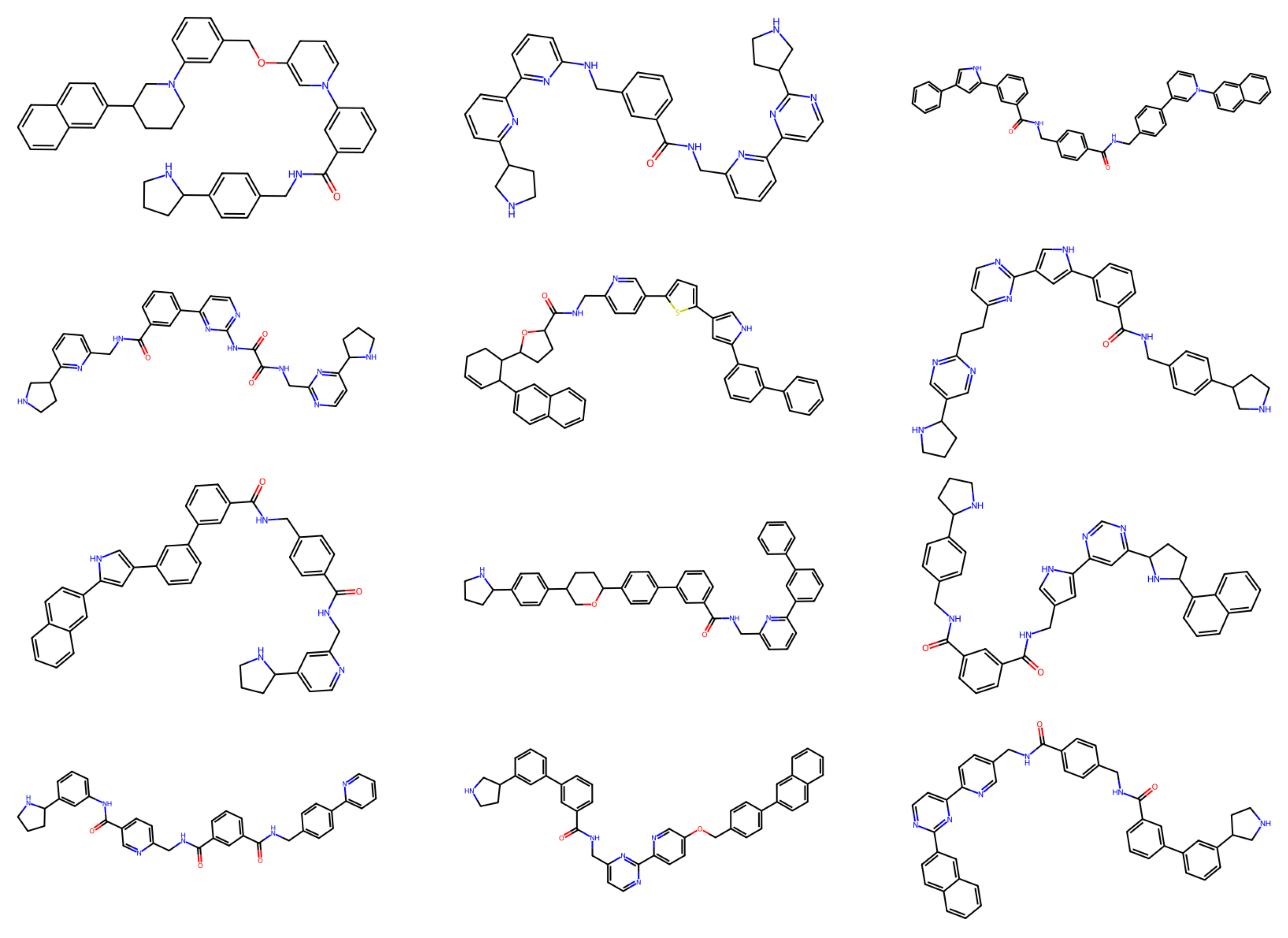}
  \caption{Top-12 molecules generated by $DGFN_{TB}$}
\end{figure}



\end{document}